\documentclass[runningheads,a4paper]{llncs}

\usepackage{amssymb}
\usepackage{graphicx}
\usepackage{url}

\usepackage{tabularx}
\usepackage{graphicx}

\urldef{\mailsa}\path|dominic.sanderson@brunel.ac.uk|
\newcommand{\keywords}[1]{\par\addvspace\baselineskip
\noindent\keywordname\enspace\ignorespaces#1}

\begin{document}

\mainmatter  

\title{Maintaining Performance with Less Data: Understanding Useful Data}

\titlerunning{Maintaining Performance with Less Data: Understanding Useful Data}

%
%
\author{Dominic Sanderson \and Tatiana Kalganova}
\authorrunning{Dominic Sanderson, Tatiana Kalganova}

\institute{Brunel University London\\
London, United Kingdom\\
\mailsa\\
\url{0000-0002-1339-143X}
}

%
%

\toctitle{Lecture Notes in Computer Science}
\tocauthor{Authors' Instructions}
\maketitle

\begin{abstract}
  As Deep Learning tasks become more popular, 
  their computational complexity increases, leading to more intricate algorithms and models which have longer runtimes 
  and require more input data. The result is a greater cost on time, hardware, and environmental resources. 
  We propose a novel method for training a neural network for image classification to reduce input data dynamically, 
  in order to reduce the costs of training a neural network model. 
  By using data reduction techniques, we reduce the amount of work performed 
  and therefore the environmental impact of AI techniques, and with dynamic data reduction we show that 
  accuracy may be maintained while reducing runtime by up to 50\%, and reducing carbon emission 
  proportionally, and gives a way forward for the better implementation of useful data. 
\end{abstract}

\section{Introduction}
When creating a deep learning solution, there are two main factors that determine its success: 
first, the model used, and second, the data used to train with. The model, often a Neural Network, may be programmed, 
optimised, and experimented with to reach an acceptable performance, but the data used to train it 
is argued to be the determining factor \cite{motamedi2021data}.
If inappropriate data is used to train and test a model,
the model will perform poorly when deployed in real-world situations and can be regarded as 
a bottleneck to the system's performance \cite{roh2019survey}. As such, consideration must be given to ensure not only an appropriate 
quantity of data is available, but also data of suitable quality \cite{roh2019survey}. 
It has been shown that increasing the data available to a model improves average accuracy \cite{linjordet2019impact}. 
Because of this, when creating new models for different tasks, researchers and data scientists 
alike will give as much data as possible to train the model, wherever possible, without considering if it is 
truly necessary. This direction is moving rapidly to a state that is environmentally unsustainable and technically 
unachievable for those outside of big tech companies \cite{thompson2020computational}.

In this paper, we analyze three novel methods to dynamically allocate the data used to train a 
neural network model, for Image Classification tasks.

The article is organized as follows: In Section \ref{prev}, we review related methods to reduce and increase 
the data used for training, as well as existing methods to use dynamically changing training data.
In Section \ref{meth}, we describe the methods and environments used for our experimentation, in Section \ref{expe} 
we describe the experimentals techniques used in our investigation of dynamic data use. In Section \ref{disc} 
we discuss the results of the experimentataion, and conclude our work in Section \ref{conc}.

\section{Previous Work}\label{prev}

\subsection{Increasing data use}\label{idu}
It is well established that a Deep Learning model will have increased performance if provided with more training data. 
Acquiring additional data may be achieved by manually gathering data, a process commonly performed either as an 
individual or part of a group/company, by using freely available data licenced as open-source, or via methods such as 
crowdfunding \cite{roh2019survey}\cite{thompson2020computational}. Alternatively, data augmentation may be used. Data augmentation is a powerful technique that 
increases the amount of data available to a model \cite{cubuk2020randaugment}. Instead of collecting more data samples, 
currently existing data samples may be modified to give new, unique datapoints for training. 
This is preferable over manual data collection, as annotated data is relatively scarce and can be difficult
to obtain, as well as requiring expert knowledge in order to label and validate the data
\cite{roh2019survey}\cite{figueroa2012predicting}.
Table \ref{tab:topAug} shows the current State of the Art models for image classification across six datasets. 
Each model uses some form of data augmentation to artificially increase the amount and variety of data available 
for training, which gives the model the improved accuracy that puts the model at the top of their leaderboard.

\begin{table}[h]
  \centering
  \caption{Top performing Image Classification models and their data augmentation methods}
  \makebox[\textwidth][c]{
  \begin{tabular}{crcrc}
    \hline
      Dataset          & \begin{tabular}[c]{@{}c@{}}No. training \\ images\end{tabular} & Model                                                                                   & Top Accuracy & Extra data/Augmentation used                                                 \\ \hline
      MNIST            & 60,000                                                        & \begin{tabular}[c]{@{}c@{}}Homogeneous \\ ensemble with \\ Simple CNN \cite{an2020ensemble}\end{tabular} & 99.91\%      & Random rotation/translation                                   \\
      CIFAR-10         & 50,000                                                        & ViT-H/14 \cite{dosovitskiy2020image}                                                                  & 99.50\%      & Random horizontal flip, square crop                           \\
      CIFAR-100        & 50,000                                                        & EffNet-L2 (SAM) \cite{foret2020sharpness}                                                             & 96.08\%      & \begin{tabular}[c]{@{}c@{}}Horizontal flip, padding,   random crop, \\AutoAugment \cite{cubuk2019autoaugment} \end{tabular}  \\
      SmallNorb        & 24,300                                                        & Efficient-CapsNet \cite{mazzia2021efficient}                                                          & 98.77\%      & \begin{tabular}[c]{@{}c@{}}Downsampling, random 32x32 crop, \\ random brightness\end{tabular}  \\
      FlowerNet-102    & 1,020                                                         & CCT-14/7x2 \cite{wu2021cvt}                                                                           & 99.72\%      & RandAugment \cite{cubuk2020randaugment}, AutoAugment  \cite{cubuk2019autoaugment}              \\
      iNaturalist 2018 & 675,170                                                       & MetaFormer \cite{diao2022metaformer}                                                                  & 88.70\%      & RandAugment \cite{cubuk2020randaugment}, AutoAugment \cite{cubuk2019autoaugment}               \\  \hline
      \end{tabular}
  }
      \label{tab:topAug}
\end{table}

\subsection{Reducing data use}\label{rdu}
While it is popular to use methods of increasing data, there are uses for data reduction techniques. 
Such practises are uncommon, as basic linear random data exclusion causes an exponential decrease in system performance \cite{cho2015much}. 
Not only is the average accuracy decreased, but the standard deviation of accuracy increases, 
showing that the training outputs are inconsistent and difficult to verify \cite{cho2015much}. 
Contrary to this, some uses of data reduction techniques give an increase to a Neural Network model's performance. 
In fields where labelled data is sparse, for example medical imaging, the data often suffers from class imbalance. 
In such cases, undersampling techniques may be employed to reduce the amount of data of the majority class, 
alleviating the effects of class imbalance \cite{liu2008exploratory}. Less success is to be gained with 
random exclusion, but methods such as Cluster Centroids \cite{rahman2013cluster} and 
Tomek Links \cite{tomek1976generalization} select the most appropriate data to remove; 
they are often used with oversampling techniques to give an increase in model performance \cite{rahman2013addressing}.
Undersampling is a technique used primarily to resolve class imbalance and is therefore unsuitable for 
uniformly balanced datasets. However, for balanced datasets, methods exist for identifying the 
'usefulness' of data, which allows less useful data to be removed from training. 
Dimensionality reduction with algorithms such as Uniform Manifold Approximation Projection (UMAP) \cite{mcinnes2018umap} 
allows a dataset to be observable in as little as 2 or 3 dimensions. By applying the UMAP algorithm to a dataset, 
the centroid of each class (where all its datapoints revolve) can be calculated, and data may then be removed 
based on its distance from its centroid \cite{byerly2022towards}. Surprisingly, by removing data closest to its class's centroid, 
accuracy is increased, despite fewer training evaluations being performed \cite{byerly2022towards}.

\subsection{Variable data use}\label{vdu}
Varying data usage is the process of using different data throughout training. 
It is a novel concept; the convention is that a model is trained with the same data over multiple iterations, 
to allow the model's weights and biases to converge to a stable state. It would seem contradictory to add a change 
to a system that would disturb this convergence; however, there are cases where its use has improved performance, 
discussed below.

Real-time or online augmentation creates unique data each epoch. 
Compared to offline augmentation, where the data is augmented once (before the model is run), 
online augmentation performs augmentation before each epoch during training \cite{shorten2019survey}. 
Augmentation is commonly used to combat issues such as overfitting and shortage of data \cite{rai2021real}, 
but for datasets with well-balanced data it is also used to achieve higher levels of generalisability \cite{byerl0y221no}.
Transfer learning is used to improve a model from one domain by transferring information from a related domain \cite{weiss2016survey}. 
It may also be used to transfer knowledge within the same domain, to reinforce the model when new data becomes 
available \cite{chen2019transfer}. 

Foremost, transfer learning is a technique used to reduce the amount of work needed to train a model \cite{chen2019transfer}; 
a model may be trained with one set of data, and the outputs recorded; then, when more data is made available, 
training can recommence, but with the previous outputs used as a starting point. This means that the model 
need not be retrained with the original data each time more data is added. In a way, this is varying data usage; 
throughout the course of training, different data is used. The unique property of transfer learning is that 
when a model is introduced to new training data, the old training data is not also used. This can often cause 
catastrophic forgetting \cite{parisi2019continual}.
There has been no investigation of using less data at the start of training, and increasing data as training progresses.
Unlike Transfer Learning, we have all the data needed at the start of training, but we choose to withhold some of it 
as training is performed. This is to show the effect of varying data use in a controlled manner. 

The experiments described in this paper investigate some extreme cases of reducing data use at earlier stages, 
and show the effect of this on model performance. Our contributions are as follows:
\begin{itemize}
    \item Three novel methods of dynamically introducing data to a model are described, which each reduce the amount of evaluations performed:
    \begin{itemize}
        \item Data Step
        \item Data Increment
        \item Data Cut
    \end{itemize}
    \item Testing on three datasets is shown, demonstrating the positive and negative effects of these new methods 
    on network accuracy and runtime.
    \item Evidence is given that these methods may be used reduce the resources required for training 
    while maintaining or improving on the performance of the training output with a runtime reduction of over 50\%.
\end{itemize}

\section{Methods}\label{meth}
\begin{figure}[htbp]
    \includegraphics[width=\textwidth]{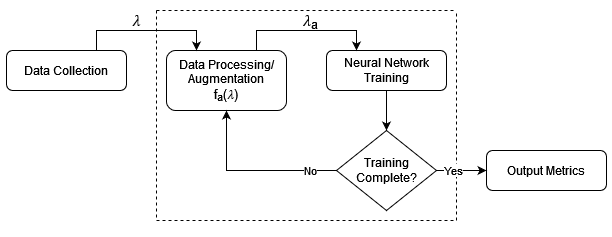}
    \centering    
    \caption{Common model structure with Online Augmentation.
    $\lambda=$ full dataset, \\ $\lambda_a=$ augmented dataset. 
    $f_a()=$ augmentation function. 
    }
    \label{figStruct1}
\end{figure}

\begin{figure}[htp]
  \includegraphics[width=\textwidth]{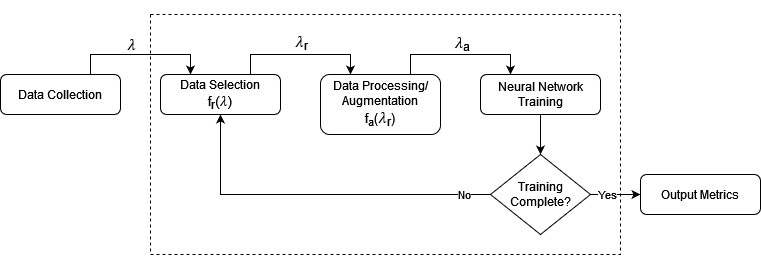}    
  \centering
  \caption{Proposed model structure with Dynamic Data Selection. 
  $\lambda=$ full dataset, $\lambda_r=$ reduced dataset, $\lambda_a=$ augmented dataset. 
  $f_r()=$ reduction function, $f_a()=$ augmentation function
  }
  \label{figStruct2}
\end{figure}

Traditionally, a deep learning task has three stages: Data Collection \cite{roh2019survey}, Data Processing \cite{kong2022efficient} \cite{kuo2019data}, 
and the training of the Neural Network Model. These steps are most often performed sequentially, but are sometimes 
somewhat interlinked; for example, the Data Collection may be real-time image or text capture, 
or the model may use Online Augmentation, where data is augmented differently each training loop. 
Figure \ref{figStruct1} shows the data flow of an Image Classification model with Online Augmentation. After each training loop, 
the model returns to the Augmentation stage. 
Figure \ref{figStruct2} shows the model structure proposed in this paper, with a data selection stage. 
This allows the model to select data dynamically, at each loop of training. This in turn enables the various 
data reduction techniques described further in this paper.

\newpage
This paper introduces and investigates three sets of experiments where, each epoch, data is selected 
for training during the data selection stage. The result is a model that does not use all data for each epoch. 
These methods are: 
\begin{itemize}
    \item Data Step
    \item Data Increment
    \item Data Cut
\end{itemize}

\subsection{Datasets}
Three datasets were used for experimentation: the handwritten digits dataset MNIST \cite{lecun1998mnist},
CIFAR10 \cite{krizhevsky2009learning}, and the model toy dataset smallNORB \cite{huang2009small}. 
Table \ref{tab:usedData} shows the properties of each dataset.

Multiple variations of experiments were performed, to observe the effect of varying amounts of 
data reduction on model performance. As such, datasets with small image sizes are ideal, 
as they required less time to train. 

\begin{table}[htbp]
    \centering
    \caption{Used dataset properties}
    \label{tab:usedData}
    \begin{tabular}{crrcc}
        \hline
        Dataset   & \begin{tabular}[c]{@{}c@{}}\# Training\\ images\end{tabular} & \begin{tabular}[c]{@{}c@{}}\# Test\\ images\end{tabular} & \begin{tabular}[c]{@{}c@{}}Image size\\ (pixels)\end{tabular} & \begin{tabular}[c]{@{}c@{}}\# Colour\\ channels\end{tabular} \\ \hline
        MNIST     & 60,000                                                       & 10,000                                                   & 28x28                                                         & 1                                                            \\
        CIFAR-10  & 50,000                                                       & 10,000                                                   & 32x32                                                         & 3                                                            \\
        smallNorb & 24,300                                                       & 24,300                                                   & 96x96                                                         & 1                                                            \\ \hline   
    \end{tabular}
\end{table}

\subsection{Hardware}
Multiple hardware were used to perform training, to make the most of resources available for this research. 
Training for each dataset was carried out on only one machine, to ensure metrics were consistent 
across a single dataset. The average CO2 emission for each GPU is also given, 
which is used for emission reduction calculations \cite{lacoste2019quantifying}. 
For training on MNIST:
\begin{itemize}
    \item CPU: AMD Ryzen 9 3950X 16-Core
    \item GPU: nVidia GeForce RTX 2070 32GB
    \item RAM: 64GB
    \item CO2 emission per hour: 0.066kg
\end{itemize}
For training on CIFAR-10:
\begin{itemize}
    \item CPU: Intel i7 10700
    \item GPU: RTX 4000
    \item RAM: 64GB
    \item CO2 emission per hour: 0.0922kg
\end{itemize}
For training on smallNorb:
\begin{itemize}
    \item CPU: AMD Ryzen 9 3950X 16-Core
    \item GPU: nVidia GeForce RTX 2070 32GB
    \item RAM: 64GB
    \item CO2 emission per hour: 0.066kg
\end{itemize}

\subsection{Performance Metrics}
These experiments are to observe the effect of reducing data used each epoch of training. 
To quantify this, the results of the experiments show the total number of evaluations performed; 
each evaluation is a single image used for training. The runtime of each experiment is also recorded. 
The successfulness of the model is quantified with the top1 accuracy achieved. 
The runtime and accuracy for each experiment is directly compared with the baseline for each dataset, 
to show the increase or decrease in performance in accuracy and runtime.
As the model is otherwise unchanged when run with data reduction techniques, the CO2 reduction of each 
experiment is directly correlated to the runtime. Therefore, CO2 reduction is equal to the runtime reduction.

\subsection{Network Architecture}
The model used for testing is a Convolutional Neural Network with a Simple Monolithic Architecture. 
It uses 9 Convolutional Layers, and a Primary and Secondary Capsule layer. The Capsule layer uses 
Homogeneous Vector Capsules, which replace the fully connected layer. The model is based on \cite{byerl0y221no}, 
which shows relatively high performance despite its few network layers.
Tests were run for 300 epochs with a batch size of 120. Optimisation was performed using the 
Adam optimiser with an initial learning rate of 0.999, with an exponential decay rate of 0.005 per epoch.
These values provide consistent initial settings across all experiments for all datasets.

The datapoints to exclude for each of the methods were selected at random. The experiments were performed to 
gather an understanding of the effect of the novel dynamic data reduction methods with the simplest method
of data exclusion, namely random data exclusion.

\section{Experimentation}\label{expe}
\subsection{Benchmark}

\begin{table}[ht]
    \centering
    \caption{Benchmark results for MNIST, CIFAR-10 and smallNorb datasets}
    \label{tab:bench}
    \makebox[\textwidth][c]{
    \begin{tabular}{crrrccccc}
        \hline
        Dataset   & \begin{tabular}[c]{@{}c@{}}Dataset \\ size\end{tabular} & \begin{tabular}[c]{@{}c@{}}\#Training \\ epochs\end{tabular} & \begin{tabular}[c]{@{}c@{}}\#Total Image \\ Evaluations\end{tabular} & \begin{tabular}[c]{@{}c@{}}Runtime \\ (hours)\end{tabular} & \begin{tabular}[c]{@{}c@{}}CO2 \\emission \\ (kg)\end{tabular} & \multicolumn{3}{c}{Test Accuracy}        \\
                  &              &                                                              &                                                                      &                                                            &                                                              & Best     & Average  & \begin{tabular}[c]{@{}c@{}}Standard \\ Deviation\end{tabular}\\ \hline
        MNIST     & 50,000       & 300                                                          & 18,000,000                                                           & 02:01                                                      & 0.134                                                        & 99.719\% & 99.701\% & 0.011\%            \\
        CIFAR-10  & 49,920       & 300                                                          & 14,976,000                                                           & 02:54                                                      & 0.268                                                        & 88.825\% & 88.689\% & 0.096\%            \\
        smallNorb & 24,240       & 300                                                          & 7,272,000                                                            & 01:27                                                      & 0.097                                                        & 93.152\% & 92.984\% & 0.135\%            \\ \hline
    \end{tabular}
    }
    \end{table}

Table \ref{tab:bench} shows the benchmark runtime and accuracy for MNIST, CIFAR-10 and smallNorb. 
These results will be used to compare the data reduction experiments, to show the reduction in runtime 
and effect on performance. The number of total image evaluations is the number of training images evaluated 
multiplied by the number of training epochs.  

\subsection{Data Step}
The simplest method of dynamically selecting data is to `step up' the data usage at a given point 
during training. A fraction of the dataset is used for a given number of training loops, after which, 
the full dataset is used for training. This splits the training process into two sections; Section one (S1), 
which uses less data, and Section two (S2), which uses the full dataset. Below are definitions for the 
two sections:
\begin{itemize}
    \item The number of epochs run in section 1: $S_1^E$
    \item The amount of data used in section 1: $S_1^D$
    \item The number of epochs run in section 2: $S_2^E$
    \item The amount of data used in section 2: $S_2^D$
\end{itemize}	
With these definitions we can derive the formula for the total number of evaluations performed:
$$n_o\; total\; evaluations=S_1^D*S_1^E+S_2^D*S_2^E$$

This data split causes a `step up' in data usage, and model will only train with a fraction of the data for some time. 
Figure \ref{figStep1}, Figure \ref{figStep2} and Figure \ref{figStep3} show how this split is applied 
between these sections.
The hypothesis is that due to less data being processed in section S1, there will be a reduction in runtime. 
After the step, the full dataset is used; this is to ensure that all features are made available at 
some point in training, although not at every epoch. This is to aid the reduction of overfitting, 
which is common when too little data is used.
Experimentation for the Data Step method is split into three parts, with three experiments each.

\subsubsection{Starting with 25\% of the dataset}
Three experiments were performed, and each experiment is comprised of two sections. 
Each section corresponds to the amount of data used in a single epoch. Epochs in in $S_1^E$ use $S_1^D$, 
and epochs in $S_1^E$ use the full dataset. Each of the experiments switch between these Sections at a 
different epoch, shown in Table \ref{tab:step1dist}. The values of $S^D$ vary between datasets as they each have
a different amount of training images.

\begin{table}[htbp]
    \centering
    \caption{Part 1 Initial data and epoch distribution}
    \label{tab:step1dist}
    \begin{tabular}{ccccc}
        \hline
    Experiment & $S_1^D$ & $S_2^D$ & $S_1^E$ & $S_2^E$ \\  \hline
    E1         & 25\%    & 100\%   & 25\%    & 75\%    \\
    E2         & 25\%    & 100\%   & 50\%    & 50\%    \\
    E3         & 25\%    & 100\%   & 75\%    & 25\%    \\  \hline
    \end{tabular}
\end{table}

\begin{figure}[htbp]
    \includegraphics[width=2.5in]{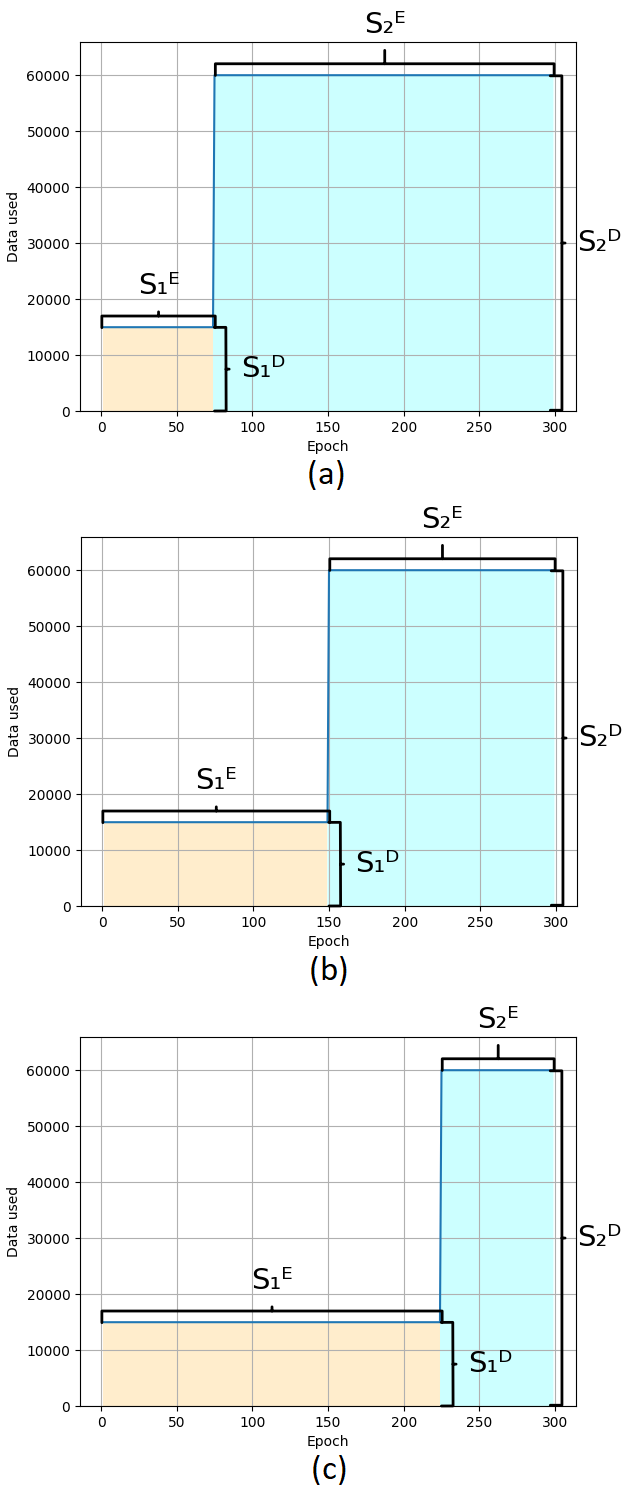}    
    \centering
    \caption{ Data Step Experiments starting with 25\% data. 
    (a) Step at 25\% through training, (b) Step at 50\% through training, (c) Step at 75\% through training     
    }
    \label{figStep1}
\end{figure}

Figure \ref{figStep1} shows graphs of the data use, for the MNIST dataset with 60,000 training images. 
The total area under the line represents the total number of evaluations. 

\begin{table}[ht]
    \centering
    \caption{Data Step part 1 experimental results}
    \label{tab:step1results}
    \makebox[\textwidth][c]{
    \begin{tabular}{crcrrrrrr}
        \hline
        Test type & \begin{tabular}[c]{@{}c@{}}\# Image \\ Evaluations\end{tabular} & \begin{tabular}[c]{@{}c@{}}Runtime \\(hours)\end{tabular} & \multicolumn{3}{c}{Test Accuracy}                                                                    & \multicolumn{3}{c}{Percentage difference}                                                                                                 \\
                  &                      &                 & Best              & Average           & \begin{tabular}[c]{@{}c@{}}Standard\\ deviation\end{tabular} & Runtime            & \begin{tabular}[c]{@{}c@{}}Best\\ Accuracy\end{tabular} & \begin{tabular}[c]{@{}c@{}}Average\\ Accuracy\end{tabular} \\ \hline
        \multicolumn{9}{c}{MNIST}                                                                                                                                                                                                                                                                             \\ \hline
        Benchmark & 18,000,000           & 02:01           & 99.719\%          & 99.701\%          & 1.10E-04                                                     &                    &                                                         &                                                            \\
        E1        & 14,678,880           & 01:43           & 99.739\%          & 99.711\%          & 2.79E-04                                                     & -15.148\%          & 0.020\%                                                 & 0.010\%                                                    \\
        E2        & 11,312,880           & 01:22           & \textbf{99.749\%} & \textbf{99.725\%} & \textbf{1.82E-04}                                            & -32.627\%          & \textbf{0.030\%}                                        & \textbf{0.024\%}                                           \\
        E3        & 7,946,880            & 00:58           & \textbf{99.749\%} & 99.709\%          & 2.35E-04                                                     & \textbf{-51.742\%} & \textbf{0.030\%}                                        & 0.008\%                                                    \\ \hline
        \multicolumn{9}{c}{CIFAR-10}                                                                                                                                                                                                                                                                          \\ \hline
        Benchmark & 14,976,000           & 02:54           & 88.825\%          & 88.689\%          & 9.59E-04                                                     &                    &                                                         &                                                            \\
        E1        & 12,214,320           & 02:25           & \textbf{88.675\%} & \textbf{88.239\%} & 2.78E-03                                                     & -16.467\%          & -0.170\%                                                & \textbf{-0.507\%}                                          \\
        E2        & 9,415,320            & 01:54           & 87.801\%          & 87.677\%          & \textbf{1.49E-03}                                            & -34.293\%          & \textbf{-1.153\%}                                       & -1.141\%                                                   \\
        E3        & 6,616,320            & 01:22           & 86.878\%          & 86.608\%          & 2.55E-03                                                     & \textbf{-52.555\%} & -2.193\%                                                & -2.346\%                                                   \\ \hline
        \multicolumn{9}{c}{smallNORB}                                                                                                                                                                                                                                                                         \\ \hline
        Benchmark & 7,272,000            & 01:27           & 93.152\%          & 92.984\%          & 1.35E-03                                                     &                    &                                                         &                                                            \\
        E1        & 5,931,120            & 01:15           & \textbf{93.172\%} & \textbf{92.944\%} & \textbf{2.22E-03}                                            & -13.922\%          & \textbf{0.022\%}                                        & \textbf{-0.043\%}                                          \\
        E2        & 4,572,120            & 01:04           & 93.003\%          & 92.588\%          & 3.70E-03                                                     & -26.674\%          & -0.159\%                                                & -0.425\%                                                   \\
        E3        & 3,213,120            & 00:51           & 92.628\%          & 92.089\%          & 4.59E-03                                                     & \textbf{-41.191\%} & -0.562\%                                                & -0.962\%                                                   \\ \hline
    \end{tabular}
    }
\end{table}

This first phase of experiments sees the largest decrease in evaluations performed compared to each benchmark. 
In particular, experiment E3 has the largest decrease out of all the Data Step experiments. As such, 
if it is assumed that modifying number of evaluations performed would affect the outputs of the program, 
then these experiments would differ the most from the benchmarks.  
The results of this first phase of experiments have shown that when fewer evaluations are performed, 
runtime is reduced. This was expected. More importantly is the effect this has the accuracy:
\begin{itemize}
    \item With the MNIST dataset, there has been a slight increase in average and best accuracy: E1, E2 and E3 show
    an average accuracy increase of 0.01\%, 0.024\% and 0.008\% respectively. 
    \item For the CIFAR-10 dataset, there is a much more noticeable decrease in accuracies. E1, E2 and E3 show 
    average accuracy decrease of 0.507\%, 1.141\% and 2.346\% respectively.
    \item For the smallNorb dataset, there is also a decrease in accuracy: E1, E2 and E3 show 
    average accuracy decrease of 0.043\%, 0.425\% and 0.962\% respectively.
\end{itemize}
Based on previous research on reducing data used for training, the results on CIFAR-10 and smallNorb datasets
is perhaps to be expected. It is curious that the MNIST shows an increase in performance, albeit a slight one. 
What is interesting still is that for both MNIST and smallNorb datasets, the standard deviation of their 
accuracies is greater than their benchmarks, but for CIFAR-10 the standard deviation is lower. 
There seems to be no pattern, and all three datasets show unique characteristics.

\subsubsection{Starting with 50\% of the dataset}
\begin{figure}[htbp]
  \includegraphics[width=2.5in]{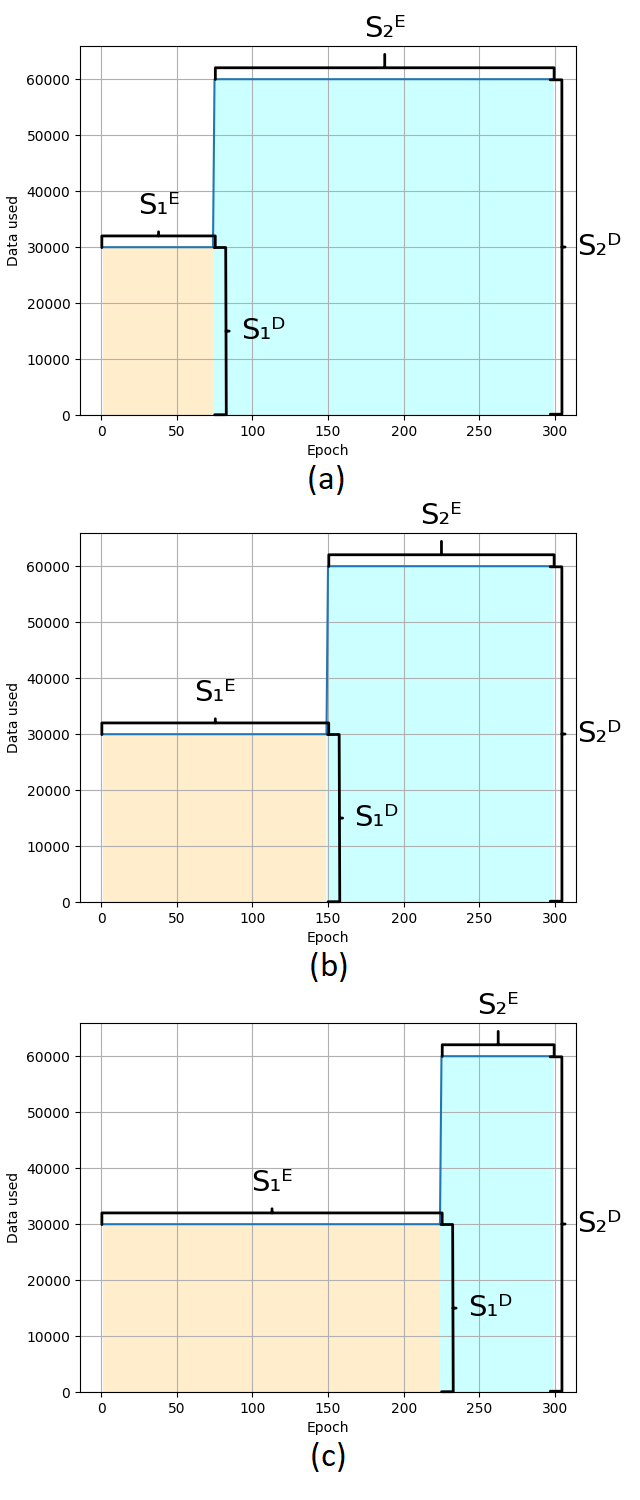}    
  \centering
  \caption{ Data Step Experiments starting with 50\% data.
  (a) Step at 25\% through training, (b) Step at 50\% through training, (c) Step at 75\% through training     
  }
  \label{figStep2}
\end{figure}

The next three experiments under the Data Step method follow the same form as in part one, 
but the initial data $S_1^D$ is 50\% of the full dataset. This is shown in Table \ref{tab:step2dist}, 
and Figure \ref{figStep2} shows graphs that
demonstrate this with the MNIST dataset. This set of experiments is to observe the effect of 
a different amount of data reduction Experimental results are shown in Table \ref{tab:step2results}.

 \begin{table}[htbp]
    \centering
    \caption{Part 2 Initial data and epoch distribution}
    \label{tab:step2dist}
    \begin{tabular}{ccccc}
        \hline
    Experiment & $S_1^D$ & $S_2^D$ & $S_1^E$ & $S_2^E$ \\  \hline
    E1         & 50\%    & 100\%   & 25\%    & 75\%    \\
    E2         & 50\%    & 100\%   & 50\%    & 50\%    \\
    E3         & 50\%    & 100\%   & 75\%    & 25\%    \\  \hline
    \end{tabular}
\end{table}

\begin{table}[]
  \centering
  \caption{Data Step part 2 experimentation results}
  \label{tab:step2results}
  \makebox[\textwidth][c]{
    \begin{tabular}{crcrrrrrr}
        \hline
        Test type & \begin{tabular}[c]{@{}c@{}}\# Image \\ Evaluations\end{tabular} & \begin{tabular}[c]{@{}c@{}}Runtime \\(hours)\end{tabular} & \multicolumn{3}{c}{Test Accuracy}                                                                    & \multicolumn{3}{c}{Percentage difference}                                                                                                 \\
                &                      &                 & Best              & Average           & \begin{tabular}[c]{@{}c@{}}Standard\\ deviation\end{tabular} & Runtime            & \begin{tabular}[c]{@{}c@{}}Best\\ Accuracy\end{tabular} & \begin{tabular}[c]{@{}c@{}}Average\\ Accuracy\end{tabular} \\ \hline
      \multicolumn{9}{c}{MNIST}                                                                                                                                                                                                                                                                             \\ \hline
      Benchmark & 18,000,000           & 02:01           & 99.719\%          & 99.701\%          & 1.10E-04                                                     &                    &                                                         &                                                            \\
      E4        & 15,788,880           & 01:53           & \textbf{99.759\%} & \textbf{99.715\%} & 2.52E-04                                                     & -6.638\%           & \textbf{0.040\%}                                        & \textbf{0.014\%}                                           \\
      E5        & 13,547,880           & 01:39           & 99.719\%          & 99.699\%          & \textbf{1.59E-04}                                            & -18.038\%          & 0.000\%                                                 & -0.002\%                                                   \\
      E6        & 11,306,880           & 01:24           & 99.729\%          & 99.695\%          & 2.98E-04                                                     & \textbf{-30.597\%} & 0.010\%                                                 & -0.006\%                                                   \\ \hline
      \multicolumn{9}{c}{CIFAR-10}                                                                                                                                                                                                                                                                          \\ \hline
      Benchmark & 14,976,000           & 02:54           & 88.825\%          & 88.689\%          & 9.59E-04                                                     &                    &                                                         &                                                            \\
      E4        & 13,137,840           & 02:35           & \textbf{88.464\%} & \textbf{88.283\%} & \textbf{1.78E-03}                                            & -10.458\%          & \textbf{-0.407\%}                                       & \textbf{-0.457\%}                                          \\
      E5        & 11,274,840           & 02:14           & 88.434\%          & 87.998\%          & 3.01E-03                                                     & -22.581\%          & -0.441\%                                                & -0.779\%                                                   \\
      E6        & 9,411,840            & 01:55           & 88.243\%          & 87.711\%          & 3.00E-03                                                     & \textbf{-33.831\%} & -0.656\%                                                & -1.103\%                                                   \\ \hline
      \multicolumn{9}{c}{smallNORB}                                                                                                                                                                                                                                                                         \\ \hline
      Benchmark & 7,272,000            & 01:27           & 93.152\%          & 92.984\%          & 1.35E-03                                                     &                    &                                                         &                                                            \\
      E4        & 6,384,000            & 01:20           & \textbf{93.564\%} & \textbf{92.978\%} & 4.45E-03                                                     & -8.475\%           & \textbf{0.443\%}                                        & \textbf{-0.006\%}                                          \\
      E5        & 5,484,000            & 01:11           & 93.263\%          & 92.500\%          & 6.19E-03                                                     & -18.317\%          & 0.120\%                                                 & -0.520\%                                                   \\
      E6        & 4,584,000            & 01:04           & 93.544\%          & 92.905\%          & \textbf{4.39E-03}                                            & \textbf{-26.675\%} & 0.421\%                                                 & -0.084\%                                                   \\  \hline
  \end{tabular}
  }
\end{table}

The results of this second phase of experiments mirrors those of the previous phase of experiments. 
We observe that with fewer evaluations performed, runtime is reduced; though as more evaluations are 
performed compared to the previous phase, the runtime is decreased less. The accuracies show the following 
patterns:
\begin{itemize}
  \item With the MNIST dataset, there is a slight increase in average and best accuracy for all but one 
  experiments:  E4 shows an average accuracy increase of 0.014\%, while E5 and E6 show average accuracy 
  decrease of 0.002\% and 0.006\% respectively. 
  \item For the CIFAR-10 dataset, there is still a decrease in accuracies across all experiments, 
  but not as steep a decline as in the previous experiments.  E4, E5 and E6 show 
  average accuracy decrease of 0.457\%, 0.779\% and 1.103\% respectively. 
  \item For the smallNorb dataset, there is also an accuracy decrease, but less extreme than in the 
  previous phase. E4, E5 and E6 show average accuracy decrease of 0.006\%, 0.520\% and 0.084\% respectively.
  The average accuracy of experiment E5 is also much higher than the others, 
  and also has a higher standard deviation, which shows that the metrics are less stable.
\end{itemize}

The expectation is that as these experiments use more data than the last set, the accuracies should be improved.
However, experiments E5 and E6 for MNIST and E5 for smallNorb show worse accuracies. This may show that dynamic data
reduction in this particular way hinders the accuracy of the model, for these particular datasets.

\subsubsection{Starting with 75\% of the dataset}
\begin{figure}[htbp]
  \includegraphics[width=2.5in]{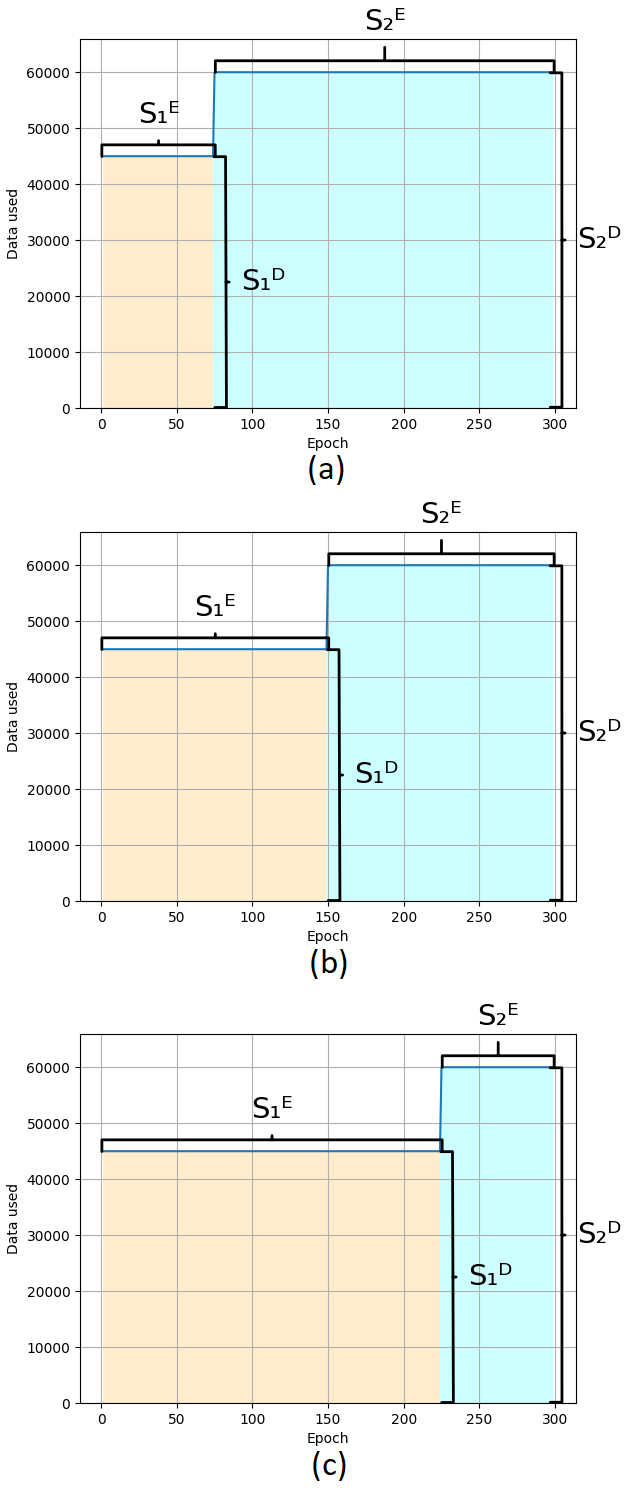}    
  \centering
  \caption{ Data Step Experiments starting with 75\% data.
  (a) Step at 25\% through training, (b) Step at 50\% through training, (c) Step at 75\% through training     
  }
  \label{figStep3}
\end{figure}

The last lot of three experiments were performed, also with two sections each. 
Epochs in $S_1^E$ use $S_1^D$, and epochs in $S_1^E$ use the full dataset $S_2^D$. In the same manner as the 
previous two lots of experiments, each of the experiments switch between these sections at a different epoch, 
shown in Table \ref{tab:step3dist}. Figure \ref{figStep3} shows graphs that demonstrate this. The aim of these experiments 
is to observe the effect of a smaller data reduction than in previous tests. 
Table \ref{tab:step3results} shows the results of the experimentation.

\begin{table}[h]
  \centering
  \caption{Part 3 Initial data and epoch distribution}
  \label{tab:step3dist}
  \begin{tabular}{ccccc}
    \hline
  Experiment & $S_1^D$ & $S_2^D$ & $S_1^E$ & $S_2^E$ \\  \hline
  E1         & 75\%    & 100\%   & 25\%    & 75\%    \\
  E2         & 75\%    & 100\%   & 50\%    & 50\%    \\
  E3         & 75\%    & 100\%   & 75\%    & 25\%    \\ \hline
  \end{tabular}
\end{table}
\newpage

\begin{table}[]
  \centering
  \caption{Data Step part 3 experimentation results}
  \label{tab:step3results}
  \makebox[\textwidth][c]{
    \begin{tabular}{crcrrrrrr}
        \hline
        Test type & \begin{tabular}[c]{@{}c@{}}\# Image \\ Evaluations\end{tabular} & \begin{tabular}[c]{@{}c@{}}Runtime \\(hours)\end{tabular} & \multicolumn{3}{c}{Test Accuracy}                                                                    & \multicolumn{3}{c}{Percentage difference}                                                                                                 \\
                &                      &                 & Best              & Average           & \begin{tabular}[c]{@{}c@{}}Standard\\ deviation\end{tabular} & Runtime            & \begin{tabular}[c]{@{}c@{}}Best\\ Accuracy\end{tabular} & \begin{tabular}[c]{@{}c@{}}Average\\ Accuracy\end{tabular} \\ \hline
      \multicolumn{9}{c}{MNIST}                                                                                                                                                                                                                                                                             \\ \hline
      Benchmark & 18,000,000           & 02:01           & 99.719\%          & 99.701\%          & 1.10E-04                                                     &                    &                                                         &                                                            \\
      E7        & 16,898,880           & 01:55           & \textbf{99.729\%} & \textbf{99.719\%} & \textbf{1.00E-04}                                            & -5.412\%           & \textbf{0.010\%}                                        & \textbf{0.018\%}                                           \\
      E8        & 15,782,880           & 01:54           & \textbf{99.729\%} & 99.717\%          & 1.31E-04                                                     & -5.713\%           & \textbf{0.010\%}                                        & 0.016\%                                                    \\
      E9        & 14,666,880           & 01:47           & \textbf{99.729\%} & 99.713\%          & 1.52E-04                                                     & \textbf{-11.948\%} & \textbf{0.010\%}                                        & 0.012\%                                                    \\ \hline
      \multicolumn{9}{c}{CIFAR-10}                                                                                                                                                                                                                                                                          \\ \hline
      Benchmark & 14,976,000           & 02:54           & 88.825\%          & 88.689\%          & 9.59E-04                                                     &                    &                                                         &                                                            \\
      E7        & 14,061,360           & 02:47           & \textbf{88.675\%} & \textbf{88.625\%} & 5.59E-04                                                     & -4.027\%           & \textbf{-0.170\%}                                       & \textbf{-0.072\%}                                          \\
      E8        & 13,134,360           & 02:36           & 88.655\%          & 88.402\%          & 2.98E-03                                                     & -9.888\%           & -0.192\%                                                & -0.324\%                                                   \\
      E9        & 12,207,360           & 02:25           & 88.353\%          & 88.239\%          & \textbf{1.14E-03}                                            & \textbf{-16.565\%} & -0.531\%                                                & -0.507\%                                                   \\ \hline
      \multicolumn{9}{c}{smallNORB}                                                                                                                                                                                                                                                                         \\ \hline
      Benchmark & 7,272,000            & 01:27           & 93.152\%          & 92.984\%          & 1.35E-03                                                     &                    &                                                         &                                                            \\
      E7        & 6,828,000            & 01:23           & \textbf{93.705\%} & \textbf{93.059\%} & 5.45E-03                                                     & -5.036\%           & \textbf{0.593\%}                                        & \textbf{0.081\%}                                           \\
      E8        & 6,378,000            & 01:19           & 93.581\%          & 92.926\%          & 5.19E-03                                                     & -9.528\%           & 0.461\%                                                 & -0.062\%                                                   \\
      E9        & 5,928,000            & 01:16           & 93.300\%          & 92.751\%          & \textbf{3.85E-03}                                            & \textbf{-13.423\%} & 0.159\%                                                 & -0.250\%                                                   \\ \hline
  \end{tabular}
  }
\end{table}

This final phase of testing uses the least data reduction, and as such, the decrease in both runtime and 
average accuracy is the smallest. The following patterns are observed:
\begin{itemize}
  \item With the MNIST dataset there is an increase in accuracy for all experiments:
  E7, E8 and E9 show an average accuracy increase of 0.018\%, 0.016\% and 0.012\% respectively.
  Experiment E7 shows a decrease in standard deviation of accuracies, compared to the benchmark. 
  \item With the CIFAR-10 dataset, the results follow the same patterns as with the results of the 
  last groups of experiments - the accuracy decrease directly correlates to the reduction in data used, 
  although this correlation is non-linear.
  E7, E8 and E9 show an average accuracy decrease of 0.072\%, 0.324\% and 0.507\% respectively.
  \item With the smallNorb dataset, 
  E7 shows an average accuracy increase of 0.081\%, and E8 and E9 show an average accuracy decrease of 
  0.062\% and 0.250\% respectively.
  There is an increase in performance with experiment E7; this is the only experiment on the smallNorb dataset 
  to have an increase in performance. The average accuracies of experiments E8 and E9 are decreased, 
  which shows how much of an effect such a slight difference in number of evaluations performed has on the model. 
\end{itemize}

\newpage
\subsection{Data Increment}
Like the methods described in the Step section, the incrementing method uses a fraction of the data at the 
start of training. At a previously defined interval, more of the data is added incrementally, causing multiple 
smaller steps in data usage throughout the training. As such, the number of sections present is equal to the 
number of increments.

\begin{figure}[ht]
    \includegraphics[width=3in]{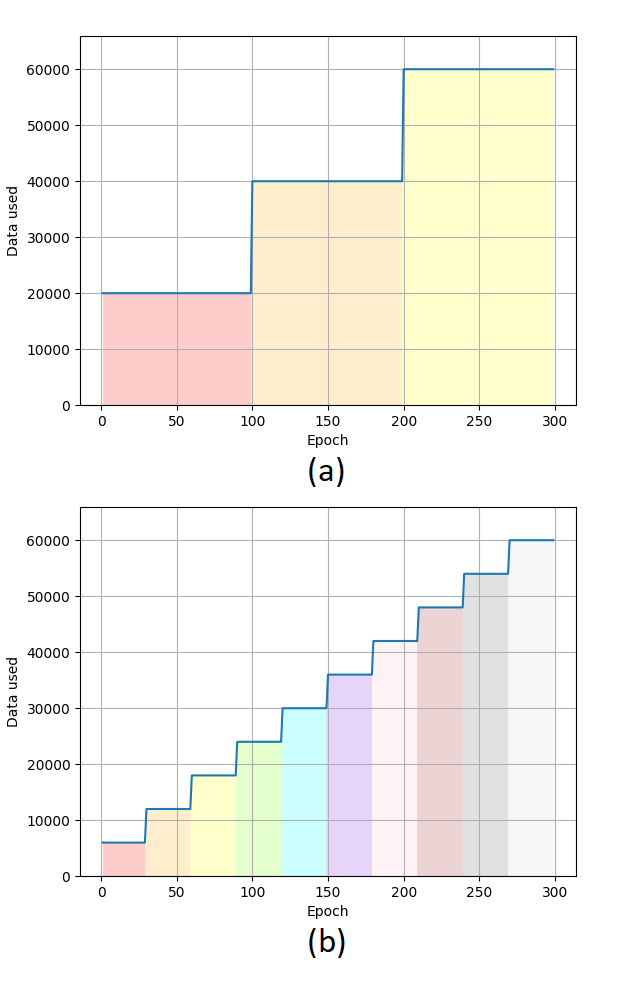}    
    \centering
    \caption{ Data Increment experiments, with
    (a) 33\% data increments. (b) 10\% data increments.       
    }
    \label{figInc}
\end{figure}

Adding the data this way will decrease the time spent training for the earlier epochs, while ensuring that 
all training data is eventually used. Figure \ref{figInc} shows example data usage graphs with varying increment 
intervals. 
Table \ref{tab:incResults} shows the results of the experimentation.

\begin{table}[]
    \centering
    \caption{Data Increment experimentation results}
    \label{tab:incResults}
    \makebox[\textwidth][c]{
        \begin{tabular}{crcrrrrrr}
            \hline
        Test type & \begin{tabular}[c]{@{}c@{}}\# Image \\ Evaluations\end{tabular} & \begin{tabular}[c]{@{}c@{}}Runtime \\(hours)\end{tabular} & \multicolumn{3}{c}{Test Accuracy}                                                                    & \multicolumn{3}{c}{Percentage difference}                                                                                                 \\
                  &                      &                 & Best              & Average           & \begin{tabular}[c]{@{}c@{}}Standard\\ deviation\end{tabular} & Runtime            & \begin{tabular}[c]{@{}c@{}}Best\\ Accuracy\end{tabular} & \begin{tabular}[c]{@{}c@{}}Average\\ Accuracy\end{tabular} \\ \hline
        \multicolumn{9}{c}{MNIST}                                                                                                                                                                                                                                                                             \\ \hline
        Benchmark & 18,000,000           & 02:01           & 99.719\%          & 99.701\%          & 1.10E-04                                                     &                    &                                                         &                                                            \\
        33\%      & 11,992,080           & 01:23           & \textbf{99.739\%} & \textbf{99.703\%} & 2.89E-04                                                     & -31.790\%          & \textbf{0.020\%}                                        & \textbf{0.000\%}                                           \\
        25\%      & 11,241,120           & 01:20           & 99.719\%          & 99.699\%          & \textbf{1.42E-04}                                            & -34.210\%          & 0.000\%                                                 & \textbf{0.000\%}                                           \\
        10\%      & 9,903,720            & 01:09           & 99.709\%          & 99.687\%          & 2.50E-04                                                     & -42.630\%          & -0.010\%                                                & -0.010\%                                                   \\
        5\%       & 9,456,720            & 01:07           & 99.729\%          & 99.701\%          & 1.93E-04                                                     & -44.820\%          & 0.010\%                                                 & \textbf{0.000\%}                                           \\
        1\%       & 9,099,120            & 01:08           & 99.709\%          & 99.677\%          & 2.18E-04                                                     & -44.090\%          & -0.010\%                                                & -0.020\%                                                   \\
        0.33\%    & 9,041,880            & \textbf{01:04}  & 99.719\%          & 99.695\%          & 1.96E-04                                                     & \textbf{-47.310\%} & 0.000\%                                                 & -0.010\%                                                   \\ \hline
        \multicolumn{9}{c}{CIFAR-10}                                                                                                                                                                                                                                                                          \\ \hline
        Benchmark & 14,976,000           & 02:54           & 88.825\%          & 88.689\%          & 9.59E-04                                                     &                    &                                                         &                                                            \\
        33\%      & 9,981,360            & 02:00           & \textbf{88.153\%} & \textbf{87.723\%} & 2.69E-03                                                     & -30.920\%          & \textbf{-0.760\%}                                       & \textbf{-1.090\%}                                          \\
        25\%      & 9,361,560            & 01:52           & 87.771\%          & 87.582\%          & 2.18E-03                                                     & -35.130\%          & -1.190\%                                                & -1.250\%                                                   \\
        10\%      & 8,253,000            & 01:42           & 87.550\%          & 87.317\%          & 1.52E-03                                                     & -41.380\%          & -1.440\%                                                & -1.550\%                                                   \\
        5\%       & 7,877,520            & 01:36           & 87.490\%          & 87.275\%          & 1.92E-03                                                     & -44.530\%          & -1.500\%                                                & -1.590\%                                                   \\
        1\%       & 7,581,360            & \textbf{01:33}  & 87.299\%          & 87.106\%          & \textbf{1.25E-03}                                            & -46.370\%          & -1.720\%                                                & -1.780\%                                                   \\
        0.33\%    & 7,531,560            & \textbf{01:33}  & 87.560\%          & 87.225\%          & 2.06E-03                                                     & \textbf{-46.560\%} & -1.420\%                                                & -1.650\%                                                   \\ \hline
        \multicolumn{9}{c}{smallNORB}                                                                                                                                                                                                                                                                         \\ \hline
        Benchmark & 7,272,000            & 01:27           & 93.152\%          & 92.984\%          & 1.35E-03                                                     &                    &                                                         &                                                            \\
        33\%      & 4,840,200            & 01:07           & \textbf{92.954\%} & \textbf{92.482\%} & 3.61E-03                                                     & -23.040\%          & \textbf{-0.210\%}                                       & \textbf{-0.540\%}                                          \\
        25\%      & 4,536,240            & 01:03           & 92.628\%          & 92.463\%          & 2.21E-03                                                     & -28.180\%          & -0.560\%                                                & -0.560\%                                                   \\
        10\%      & 3,999,840            & 01:02           & 92.768\%          & 92.373\%          & 3.51E-03                                                     & -28.850\%          & -0.410\%                                                & -0.660\%                                                   \\
        5\%       & 3,821,040            & 00:59           & 92.735\%          & 92.291\%          & 3.45E-03                                                     & -32.180\%          & -0.450\%                                                & -0.740\%                                                   \\
        1\%       & 3,676,560            & 00:57           & 92.562\%          & 92.115\%          & 4.91E-03                                                     & -34.520\%          & -0.630\%                                                & -0.930\%                                                   \\ 
        0.33\%    & 3,651,600            & \textbf{00:56}  & 92.525\%          & 92.248\%          & \textbf{2.39E-03}                                            & \textbf{-35.700\%} & -0.670\%                                                & -0.790\%                                                   \\  \hline
    \end{tabular}
    }
\end{table}

Using the data increment method causes much less data to be used throughout training. The most extreme case of this is with steps of
0.33\%, where roughly half of the evaluations are performed. This makes the data increment method a more extreme data reduction method
than the data step method.
For each dataset, the observations are as follows:
\begin{itemize}
    \item For the MNIST dataset, average accuracy is maintained with 33\% and 25\% increments, but is never improved upon the baseline.
    For both of these cases, runtime is shortened by at least 30\%.
    \item For the CIFAR-10 dataset, all average accuracies are worse by at least 1\%. Like MNIST, runtime is shortened by a minimum of 
    30\%.
    \item For the smallNorb dataset, as the dataset has much fewer training images, both the runtime and accuracy reduction are not as
    large as with other datasets. For all experiments, average accuracy is reduced by less than 1\%, while runtime is reduced by 
    between 20\% to 40\%.
\end{itemize}
Based on these results, the data increment method seems to have a more detrimental effect on network performance. It is also worth noting
that the standard deviation of average accuracies are worse than the benchmark across all experiments, meaning the networks are less
reliable at getting consistent results.
 
\newpage
\subsection{Data Cut}
\begin{figure}[htbp]
    \includegraphics[width=3in]{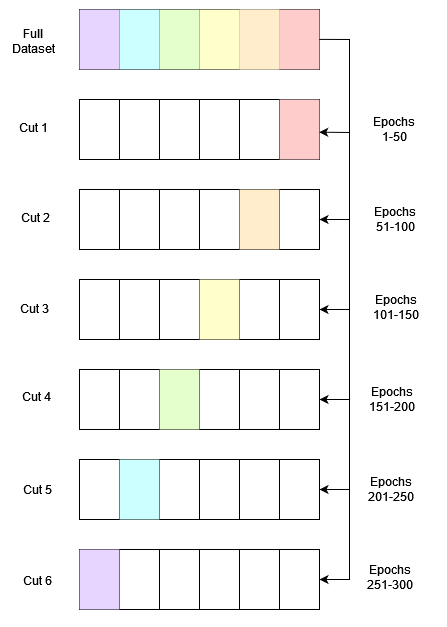}    
    \centering
    \caption{ A dataset is split into 6 sections (cuts). A single split is used for training every 
    50 epochs out of 300.      
    }
    \label{figCut}
\end{figure}

The data cut method splits a dataset into multiple cuts of equal size. 
Each cut is then used for an equal number of training epochs. The more splits a deck is cut into, 
the less data is present in each split. As such, the more splits there are, the fewer evaluations 
will be performed. Figure \ref{figCut} shows how a dataset may be divided into 6 cuts, 
for training over 300 epochs.

This approach is similar to Transfer Learning techniques, in which a network that has already been trained on some data 
is introduced to a new set of training data. The new input data contains the same classes as it had already been trained with, 
so the network does not need to adapt to new classes. However, while previous methods in this paper introduced new training data 
to accompany previously used data, this method does not include the previously used data when new data is introduced - 
it continues training on only new data.
Normally, when introducing new data via transfer learning, a fresh network is deployed, and given the weights and biases 
of previously trained networks. The Data Cut method aims to give the network different data at different times, 
while preserving the state of the network throughout a single training cycle. 
Table \ref{tab:deckResults} shows the results of the experimentation.

\begin{table}[]
  \centering
  \caption{Data Cut experimentation results}
  \label{tab:deckResults}
  \makebox[\textwidth][c]{
    \begin{tabular}{crcrrrrrr}
        \hline
        Test type & \begin{tabular}[c]{@{}c@{}}\# Image \\ Evaluations\end{tabular} & \begin{tabular}[c]{@{}c@{}}Runtime \\(hours)\end{tabular} & \multicolumn{3}{c}{Test Accuracy}                                                                    & \multicolumn{3}{c}{Percentage difference}                                                                                                 \\
            &                      &                 & Best              & Average           & \begin{tabular}[c]{@{}c@{}}Standard\\ deviation\end{tabular} & Runtime            & \begin{tabular}[c]{@{}c@{}}Best\\ Accuracy\end{tabular} & \begin{tabular}[c]{@{}c@{}}Average\\ Accuracy\end{tabular} \\ \hline
  \multicolumn{9}{c}{MNIST}                                                                                                                                                                                                                                                                             \\ \hline
  Benchmark & 18,000,000           & 02:01           & 99.719\%          & 99.701\%          & 1.10E-04                                                     &                    &                                                         &                                                            \\
  2 Splits  & 9,000,000            & 01:07           & \textbf{99.719\%} & \textbf{99.699\%} & \textbf{1.23E-04}                                            & -44.272\%          & \textbf{0.000\%}                                        & \textbf{-0.002\%}                                          \\
  3 Splits  & 5,976,000            & 00:44           & 99.679\%          & 99.655\%          & 1.68E-04                                                     & -63.331\%          & -0.040\%                                                & -0.046\%                                                   \\
  6 Splits  & 2,988,000            & 00:25           & 99.659\%          & 99.596\%          & 4.10E-04                                                     & -78.800\%          & -0.060\%                                                & -0.105\%                                                   \\
  9 Splits  & 1,980,000            & 00:19           & 99.598\%          & 99.526\%          & 6.82E-04                                                     & -84.016\%          & -0.121\%                                                & -0.175\%                                                   \\
  12 Splits & 1,476,000            & 00:16           & 99.508\%          & 99.337\%          & 1.41E-03                                                     & -86.604\%          & -0.211\%                                                & -0.365\%                                                   \\
  15 Splits & 1,188,000            & 00:14           & 99.367\%          & 99.102\%          & 2.85E-03                                                     & -88.074\%          & -0.352\%                                                & -0.600\%                                                   \\
  18 Splits & 972,000              & 00:13           & 99.297\%          & 98.855\%          & 3.86E-03                                                     & -89.124\%          & -0.423\%                                                & -0.848\%                                                   \\
  21 Splits & 828,000              & 00:12           & 99.237\%          & 97.476\%          & 1.83E-02                                                     & -89.909\%          & -0.483\%                                                & -2.232\%                                                   \\
  24 Splits & 720,000              & 00:11           & 96.365\%          & 95.645\%          & 7.33E-03                                                     & \textbf{-90.501\%} & -3.363\%                                                & -4.068\%                                                   \\ \hline
  \multicolumn{9}{c}{CIFAR-10}                                                                                                                                                                                                                                                                          \\ \hline
  Benchmark & 14,976,000           & 02:54           & 88.825\%          & 88.689\%          & 9.59E-04                                                     &                    &                                                         &                                                            \\
  2 Piles   & 7,488,000            & 01:33           & \textbf{86.777\%} & \textbf{86.528\%} & \textbf{1.54E-03}                                            & -46.518\%          & \textbf{-2.306\%}                                       & \textbf{-2.436\%}                                          \\
  3 Splits  & 4,968,000            & 01:04           & 85.492\%          & 85.143\%          & 3.20E-03                                                     & -62.692\%          & -3.753\%                                                & -3.998\%                                                   \\
  6 Splits  & 2,484,000            & 00:37           & 82.269\%          & 81.984\%          & 3.27E-03                                                     & -78.491\%          & -7.381\%                                                & -7.560\%                                                   \\
  9 Splits  & 1,656,000            & 00:28           & 80.110\%          & 79.673\%          & 2.65E-03                                                     & -83.690\%          & -9.811\%                                                & -10.166\%                                                  \\
  12 Splits & 1,224,000            & 00:23           & 77.791\%          & 77.072\%          & 6.68E-03                                                     & -86.431\%          & -12.422\%                                               & -13.098\%                                                  \\
  15 Splits & 972,000              & 00:20           & 75.552\%          & 72.888\%          & 2.02E-02                                                     & -88.013\%          & -14.943\%                                               & -17.816\%                                                  \\
  18 Splits & 828,000              & 00:19           & 72.440\%          & 69.066\%          & 2.70E-02                                                     & -88.971\%          & -18.447\%                                               & -22.125\%                                                  \\
  21 Splits & 684,000              & 00:17           & 67.731\%          & 66.030\%          & 1.94E-02                                                     & -89.880\%          & -23.748\%                                               & -25.548\%                                                  \\
  24 Splits & 612,000              & 00:16           & 64.518\%          & 60.600\%          & 4.62E-02                                                     & \textbf{-90.370\%} & -27.365\%                                               & -31.671\%                                                  \\ \hline
  \multicolumn{9}{c}{smallNORB}                                                                                                                                                                                                                                                                         \\ \hline
  Benchmark & 7,272,000            & 01:27           & 93.152\%          & 92.984\%          & 1.35E-03                                                     &                    &                                                         &                                                            \\
  2 Splits  & 3,636,000            & 00:59           & \textbf{93.106\%} & \textbf{92.744\%} & 3.37E-03                                                     & -32.490\%          & \textbf{-0.049\%}                                       & \textbf{-0.257\%}                                          \\
  3 Splits  & 2,412,000            & 00:51           & 92.880\%          & 92.616\%          & \textbf{2.36E-03}                                            & -41.598\%          & -0.292\%                                                & -0.396\%                                                   \\
  6 Splits  & 1,188,000            & 00:42           & 92.946\%          & 92.104\%          & 5.30E-03                                                     & -52.148\%          & -0.221\%                                                & -0.946\%                                                   \\
  9 Splits  & 792,000              & 00:38           & 91.840\%          & 91.408\%          & 3.93E-03                                                     & -55.659\%          & -1.408\%                                                & -1.694\%                                                   \\
  12 Splits & 576,000              & 00:36           & 90.924\%          & 90.012\%          & 7.59E-03                                                     & -58.625\%          & -2.392\%                                                & -3.195\%                                                   \\
  15 Splits & 468,000              & 00:36           & 89.975\%          & 87.791\%          & 2.33E-02                                                     & -58.774\%          & -3.410\%                                                & -5.584\%                                                   \\
  18 Splits & 396,000              & 00:36           & 88.461\%          & 87.139\%          & 1.05E-02                                                     & -58.445\%          & -5.035\%                                                & -6.286\%                                                   \\
  21 Splits & 324,000              & 00:36           & 86.865\%          & 80.511\%          & 8.85E-02                                                     & -58.097\%          & -6.749\%                                                & -13.414\%                                                  \\
  24 Splits & 288,000              & 00:37           & 86.894\%          & 80.776\%          & 7.48E-02                                                     & \textbf{-57.390\%} & -6.718\%                                                & -13.128\%                                                  \\  \hline
  \end{tabular}
  }
\end{table}

\begin{figure}[ht]
  \includegraphics[width=3in]{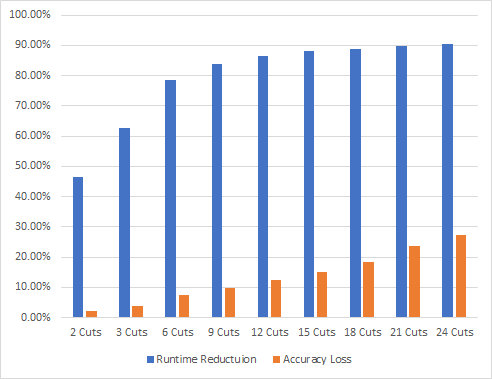}    
  \centering
  \caption{ Runtime reduction and accuracy loss with Data Cut method, on CIFAR-10 dataset     
  }
  \label{fig:cifarSplit}
\end{figure}

Of the three experimental methods described in this paper, the Data Cut method cuts the number of 
evaluations performed down the most. It becomes apparent quickly that the more splits are used, 
the more the accuracy is degraded.
Results on the MNIST dataset show that by cutting the dataset into two splits, we almost cut 
the runtime in half while maintaining the benchmark accuracy.
With 24 splits, there is a large drop in performance. This shows a `cut-off point', 
where too few evaluations are performed, and the model doesn't have enough iterations to 
allow the weights and biases to converge, causing more classifications to fail. 
Figure \ref{fig:cifarSplit} shows the percentage difference in runtime and accuracy of the CIFAR-10 dataset, 
compared to its benchmark. The results of this dataset show the biggest percentage differences. 
The graph shows that by introducing more data splits, the runtime decreases non-linearly, 
while the accuracy decreases more steadily. Therefore, the most appropriate number of data splits 
must be chosen to benefit the most from this dataset reduction technique.

\section{Discussion}\label{disc}
The runtime of the experiment correlates directly with the total number of evaluations performed. 
This is as expected, as the fewer evaluations are performed, the fewer calculations are performed by the 
neural network. 

For the Data Step method, the most notable results are those of the MNIST dataset, and E7, E8 and E9 of smallNorb, 
as they show an increase in accuracy despite the fewer evaluations performed. An future  
research direction would be to investigate why there is a performance increase in these cases, 
and how the detrimental effect on stability may be rectified. Also, further research into why CIFAR-10 
does not show any increase could be explored.

As with the Data Step method, the Data Increment method shows that reducing the number of evaluations performed decreases the 
runtime. However, despite both methods performing the same fundamental task of data reduction, they produce different results for 
roughly the same number of evaluations performed. For example, comparing experiments E2 of the data step method and data increments 
of 25\%, we can observe the accuracies when roughly the same number of evaluations are performed.

The number of evaluations for the two experiments are:
\begin{itemize}
    \item E2 for MNIST dataset uses 11,312,880 evaluations, data increments of 25\% use 11,241,120 evaluations,
    a difference of 0.634\%.
    \item E2 for CIFAR-10 dataset uses 9,415,320 evaluations, data increments of 25\% use 9,361,560 evaluations,
    a difference of 0.571\%.
    \item E2 for smallNorb dataset uses 4,572,120 evaluations, data increments of 25\% use 4,536,240 evaluations,
    a difference of 0.785\%.
\end{itemize}

However, despite the marginal difference in number of evaluations performed, the difference in accuracy between 
these two experiments are:

\begin{itemize}
    \item E2 for MNIST dataset gives an accuracy increase of 0.024\%, data increments of 25\% gives an accuracy difference of 0\%.
    \item E2 for CIFAR-10 dataset gives an accuracy decrease of 1.141\%, data increments of 25\% gives an accuracy decrease of 1.250\%.
    \item E2 for smallNorb dataset gives an accuracy decrease of 0.425\%, data increments of 25\% gives an accuracy difference of 0.560\%.
\end{itemize}

For all cases, the data increment method shows worse accuracy. While it is true that the data increment method has fewer evaluations
performed, the difference in data (less than 1\% for all datasets) amounts to less than 3 epochs of training. 
This amount is negligable as the top accuracy of a model settles at much earlier epochs. 
The results of the experimentation show that decreasing the number of evaluations with the Increment method has a more 
detrimental effect on the accuracy of the model.  

Finally, the data cut method shows the largest reduction in evaluations performed. 
Below is a comparison between experiments data increment of 0.33\% and data cut with 2 splits:
\begin{itemize}
    \item Data Increment for MNIST dataset uses 9,041,880 evaluations, data cut with 2 splits use 9,000,000 evaluations,
    a difference of 0.463\%.
    \item Data Increment for CIFAR-10 dataset uses 7,531,560 evaluations, data cut with 2 splits use 7,488,000 evaluations,
    a difference of 0.578\%.
    \item Data Increment for smallNorb dataset uses 3,651,600 evaluations, data cut with 2 splits use 3,636,000 evaluations,
    a difference of 0.427\%.
\end{itemize} 
Comparing the accuracies between these two experiments, we observe:
\begin{itemize}
    \item Data Increment for MNIST gives an accuracy decrease of 0.01\%, data cut with 2 splits gives an accuracy decrease of 0.02\%.
    \item Data Increment for CIFAR-10 gives an accuracy decrease of 1.65\%, data cut with 2 splits gives an accuracy decrease of 2.436\%.
    \item Data Increment for smallNorb gives an accuracy decrease of 0.79\%, data cut with 2 splits gives an accuracy decrease of 0.257\%.
\end{itemize} 
Accuracy for all observed data cut experiments are worse than those of the data increment method. We can conclude that,
as the data increment method is inferior to the data step method, the data cut method is least effective at maintaining accuracy.
However, if the need is to reduce the runtime by as much as possible without hindering accuracy too greatly, 
a data cut of 9 splits appears to be the best compromise between runtime and accuracy.

\section{Conclusion}\label{conc}
The results of this paper have shown that, contrary to the norm, reducing the data used for training has 
in some cases improved the performance of the model. It proves that not all data may be necessary for training, 
and in fact some data may hinder it. 

The approaches used are somewhat brutish, excluding data randomly without consideration of the value of 
the datapoints removed. Other works cited have shown algorithmic approaches to select which data to 
remove to enhance performance, which is a next step for varying data usage. Still, having shown that 
even random exclusion has improved results, it would seem that varying data use is to be explored in 
further detail.

\bibliographystyle{ieeetr}
\bibliography{References}

\begin{thebibliography}{10}

\bibitem{motamedi2021data}
M.~Motamedi, N.~Sakharnykh, and T.~Kaldewey, ``A data-centric approach for
  training deep neural networks with less data,'' {\em arXiv preprint
  arXiv:2110.03613}, 2021.

\bibitem{roh2019survey}
Y.~Roh, G.~Heo, and S.~E. Whang, ``A survey on data collection for machine
  learning: a big data-ai integration perspective,'' {\em IEEE Transactions on
  Knowledge and Data Engineering}, vol.~33, no.~4, pp.~1328--1347, 2019.

\bibitem{linjordet2019impact}
T.~Linjordet and K.~Balog, ``Impact of training dataset size on neural answer
  selection models,'' in {\em European Conference on Information Retrieval},
  pp.~828--835, Springer, 2019.

\bibitem{thompson2020computational}
N.~C. Thompson, K.~Greenewald, K.~Lee, and G.~F. Manso, ``The computational
  limits of deep learning,'' {\em arXiv preprint arXiv:2007.05558}, 2020.

\bibitem{cubuk2020randaugment}
E.~D. Cubuk, B.~Zoph, J.~Shlens, and Q.~V. Le, ``Randaugment: Practical
  automated data augmentation with a reduced search space,'' in {\em
  Proceedings of the IEEE/CVF conference on computer vision and pattern
  recognition workshops}, pp.~702--703, 2020.

\bibitem{figueroa2012predicting}
R.~L. Figueroa, Q.~Zeng-Treitler, S.~Kandula, and L.~H. Ngo, ``Predicting
  sample size required for classification performance,'' {\em BMC medical
  informatics and decision making}, vol.~12, no.~1, pp.~1--10, 2012.

\bibitem{an2020ensemble}
S.~An, M.~Lee, S.~Park, H.~Yang, and J.~So, ``An ensemble of simple
  convolutional neural network models for mnist digit recognition,'' {\em arXiv
  preprint arXiv:2008.10400}, 2020.

\bibitem{dosovitskiy2020image}
A.~Dosovitskiy, L.~Beyer, A.~Kolesnikov, D.~Weissenborn, X.~Zhai,
  T.~Unterthiner, M.~Dehghani, M.~Minderer, G.~Heigold, S.~Gelly, {\em et~al.},
  ``An image is worth 16x16 words: Transformers for image recognition at
  scale,'' {\em arXiv preprint arXiv:2010.11929}, 2020.

\bibitem{foret2020sharpness}
P.~Foret, A.~Kleiner, H.~Mobahi, and B.~Neyshabur, ``Sharpness-aware
  minimization for efficiently improving generalization,'' {\em arXiv preprint
  arXiv:2010.01412}, 2020.

\bibitem{cubuk2019autoaugment}
E.~D. Cubuk, B.~Zoph, D.~Mane, V.~Vasudevan, and Q.~V. Le, ``Autoaugment:
  Learning augmentation strategies from data,'' in {\em Proceedings of the
  IEEE/CVF Conference on Computer Vision and Pattern Recognition},
  pp.~113--123, 2019.

\bibitem{mazzia2021efficient}
V.~Mazzia, F.~Salvetti, and M.~Chiaberge, ``Efficient-capsnet: Capsule network
  with self-attention routing,'' {\em Scientific reports}, vol.~11, no.~1,
  pp.~1--13, 2021.

\bibitem{wu2021cvt}
H.~Wu, B.~Xiao, N.~Codella, M.~Liu, X.~Dai, L.~Yuan, and L.~Zhang, ``Cvt:
  Introducing convolutions to vision transformers,'' in {\em Proceedings of the
  IEEE/CVF International Conference on Computer Vision}, pp.~22--31, 2021.

\bibitem{diao2022metaformer}
Q.~Diao, Y.~Jiang, B.~Wen, J.~Sun, and Z.~Yuan, ``Metaformer: A unified meta
  framework for fine-grained recognition,'' {\em arXiv preprint
  arXiv:2203.02751}, 2022.

\bibitem{cho2015much}
J.~Cho, K.~Lee, E.~Shin, G.~Choy, and S.~Do, ``How much data is needed to train
  a medical image deep learning system to achieve necessary high accuracy?,''
  {\em arXiv preprint arXiv:1511.06348}, 2015.

\bibitem{liu2008exploratory}
X.-Y. Liu, J.~Wu, and Z.-H. Zhou, ``Exploratory undersampling for
  class-imbalance learning,'' {\em IEEE Transactions on Systems, Man, and
  Cybernetics, Part B (Cybernetics)}, vol.~39, no.~2, pp.~539--550, 2008.

\bibitem{rahman2013cluster}
M.~M. Rahman and D.~Davis, ``Cluster based under-sampling for unbalanced
  cardiovascular data,'' in {\em Proceedings of the world congress on
  engineering}, vol.~3, pp.~3--5, 2013.

\bibitem{tomek1976generalization}
I.~Tomek, ``A generalization of the k-nn rule,'' {\em IEEE Transactions on
  Systems, Man, and Cybernetics}, no.~2, pp.~121--126, 1976.

\bibitem{rahman2013addressing}
M.~M. Rahman and D.~N. Davis, ``Addressing the class imbalance problem in
  medical datasets,'' {\em International Journal of Machine Learning and
  Computing}, vol.~3, no.~2, p.~224, 2013.

\bibitem{mcinnes2018umap}
L.~McInnes, J.~Healy, and J.~Melville, ``Umap: Uniform manifold approximation
  and projection for dimension reduction,'' {\em arXiv preprint
  arXiv:1802.03426}, 2018.

\bibitem{byerly2022towards}
A.~Byerly and T.~Kalganova, ``Towards an analytical definition of sufficient
  data,'' {\em arXiv preprint arXiv:2202.03238}, 2022.

\bibitem{shorten2019survey}
C.~Shorten and T.~M. Khoshgoftaar, ``A survey on image data augmentation for
  deep learning,'' {\em Journal of big data}, vol.~6, no.~1, pp.~1--48, 2019.

\bibitem{rai2021real}
R.~Rai and D.~S. Sisodia, ``Real-time data augmentation based transfer learning
  model for breast cancer diagnosis using histopathological images,'' in {\em
  Advances in Biomedical Engineering and Technology}, pp.~473--488, Springer,
  2021.

\bibitem{byerl0y221no}
A.~Byerly, T.~Kalganova, and I.~Dear, ``No routing needed between capsules,''
  {\em Neurocomputing}, vol.~463, pp.~545--553, 2021.

\bibitem{weiss2016survey}
K.~Weiss, T.~M. Khoshgoftaar, and D.~Wang, ``A survey of transfer learning,''
  {\em Journal of Big data}, vol.~3, no.~1, pp.~1--40, 2016.

\bibitem{chen2019transfer}
L.~Chen and A.~Moschitti, ``Transfer learning for sequence labeling using
  source model and target data,'' in {\em Proceedings of the AAAI Conference on
  Artificial Intelligence}, vol.~33, pp.~6260--6267, 2019.

\bibitem{parisi2019continual}
G.~I. Parisi, R.~Kemker, J.~L. Part, C.~Kanan, and S.~Wermter, ``Continual
  lifelong learning with neural networks: A review,'' {\em Neural Networks},
  vol.~113, pp.~54--71, 2019.

\bibitem{kong2022efficient}
F.~Kong and R.~Henao, ``Efficient classification of very large images with tiny
  objects,'' in {\em Proceedings of the IEEE/CVF Conference on Computer Vision
  and Pattern Recognition}, pp.~2384--2394, 2022.

\bibitem{kuo2019data}
C.-W. Kuo, J.~D. Ashmore, D.~Huggins, and Z.~Kira, ``Data-efficient graph
  embedding learning for pcb component detection,'' in {\em 2019 IEEE Winter
  Conference on Applications of Computer Vision (WACV)}, pp.~551--560, IEEE,
  2019.

\bibitem{lecun1998mnist}
Y.~LeCun, ``The mnist database of handwritten digits,'' {\em
  http://yann.lecun.com/exdb/mnist/}, 1998.

\bibitem{krizhevsky2009learning}
A.~Krizhevsky, G.~Hinton, {\em et~al.}, ``Learning multiple layers of features
  from tiny images,'' 2009.

\bibitem{huang2009small}
F.~J. Huang and Y.~LeCun, ``The small norb dataset, v1. 0, 2005,'' 2009.

\bibitem{lacoste2019quantifying}
A.~Lacoste, A.~Luccioni, V.~Schmidt, and T.~Dandres, ``Quantifying the carbon
  emissions of machine learning,'' {\em arXiv preprint arXiv:1910.09700}, 2019.

\end{thebibliography}

\end{document}